%% file: main.tex
\documentclass[11pt]{article}

\usepackage[preprint]{acl}

\usepackage{times}
\usepackage{latexsym}

\usepackage[T1]{fontenc}
\usepackage[utf8]{inputenc}
\usepackage{microtype}
\usepackage{inconsolata}
\usepackage{graphicx}
\usepackage{amssymb}
\usepackage{amsmath}
\usepackage{subcaption}
\usepackage{booktabs}
\usepackage{array}
\usepackage{multirow}
\usepackage[inline]{enumitem}
\usepackage{xcolor}

\usepackage{standalone}
\usepackage{tikz}
\usepackage{epigraph}
\usepackage{tabularx}
\usepackage{tcolorbox}

\usetikzlibrary{shapes.geometric, arrows.meta, positioning, shadows.blur, calc}

\definecolor{neutral}{HTML}{ECF0F1}     
\definecolor{trueColor}{HTML}{00BFC4}   
\definecolor{falseColor}{HTML}{F8766D} 
\definecolor{textColor}{HTML}{2C3E50}   

\title{Traces of Social Competence in Large Language Models}

\author{
 Tom Kouwenhoven$^{\ast}$, 
 Michiel van der Meer\thanks{Equal contribution}, and 
 Max van Duijn 
 \\
 Leiden Institute of Advanced Computer Science\\Leiden University, Netherlands\\\texttt{\{t.kouwenhoven, m.t.van.der.meer, m.j.van.duijn\}@liacs.leidenuniv.nl} \\
}

\begin{document}
\maketitle
\begin{abstract}
The False Belief Test (FBT) has been the main method for assessing Theory of Mind (ToM) and related socio-cognitive competencies.
For Large Language Models (LLMs), the reliability and explanatory potential of this test have remained limited due to issues like data contamination, insufficient model details, and inconsistent controls. 
We address these issues by testing 17 open-weight models on a balanced set of 192 FBT variants \cite{trott2023do} using Bayesian Logistic regression to identify how model size and post-training affect socio-cognitive competence. 
We find that scaling model size benefits performance, but not strictly.
A cross-over effect reveals that explicating propositional attitudes (X \textit{thinks}) fundamentally alters response patterns.
Instruction tuning partially mitigates this effect, but further reasoning-oriented fine-tuning amplifies it.
In a case study analysing social reasoning ability throughout OLMo 2 training, we show that this cross-over effect emerges during pre-training, suggesting that models acquire stereotypical response patterns tied to mental-state vocabulary that can outweigh other scenario semantics. 
Finally, vector steering allows us to isolate a \emph{think} vector as the causal driver of observed FBT behaviour.
\end{abstract}

\section{Introduction}
\begin{tcolorbox}[
    colback=white,
    colframe=trueColor,
    arc=4pt,
    boxrule=0.5pt,
]
\small\textit{``Maxi and his mother store chocolate in the \textcolor{blue}{blue} cupboard in their kitchen. When Maxi leaves for the playground, his mother takes the chocolate and eats a piece. Then she puts it back into the \textcolor{green}{green} cupboard and leaves. When Maxi comes home, where will he look for the chocolate?''}
\end{tcolorbox}

\noindent
A version of this scenario was presented to children aged 3-9 by Wimmer and Perner in the early 1980s, after which none of the 3-4 year old, $57\%$ of the 4-6 year old, and $86\%$ of the 6-9 year old children in their sample pointed ``correctly'' to the blue cupboard \cite{wimmer1983beliefs}. 
This finding was interpreted as reflecting children's emerging capacity to work with the difference between one's own and somebody else's relation to the same propositional content, a form of meta-representation \cite{Pylyshyn1978when}. 
Over the ensuing decades, variants of this scenario were implemented in experiments with a wide variety of human and non-human populations, including LLMs recently.
Across all this work, the test subjects' ability to appreciate that a character may hold a False Belief has been taken as an indication of their broader social-cognitive competence, often described as their capacity for ToM \cite{apperly2010mindreaders}.

The relationship between language and the representation of false beliefs remains debated for both humans and LLMs.
Studies in child development have demonstrated correlational and causal links between the mastery of certain linguistic forms (i.a., negation, clausal complement syntax, verbs of cognition) and FBT performance, suggesting that language acquisition may scaffold meta-representational ability \cite{milligan2007language}. 
For LLMs, the debate centres on whether distributional patterns in their linguistic training data amount to ``social-world models'' that generalise beyond highly specific contexts, such as the FBT, or whether they only learn to solve such tests using superficial, context-specific heuristics. 
Over the past years, approaches for evaluating and enhancing socio-cognitive capacities in LLMs, including FBT, have proliferated \cite[see][and \autoref{background}, for an overview]{chen2025theorymindlargelanguage}.

However, common problems remain that, for LLMs, test and evaluation approaches vary in robustness.
Oftentimes, a small number of LLMs are used, which complicates the determination of the generalisability of empirical findings \cite{trott2025toward}.
Little information is released about the majority of models, making it difficult to control for leakage of test data into the training data and to systematically map the effects of differences in model size, architecture, pre-, mid- and post-training \cite{hu2025re-evaluatingToM}. 
In addition, difficulties in disentangling test formulations from underlying model representations have hampered progress on fundamental questions of mechanisms and generalisability. 
Addressing these concerns is the focus of this paper. 

Leveraging the increasing availability of open-source and open-weight model families, we provide such a systematic comparison for a total of 17 model variants based on computed probabilities \cite{pimentel-meister-2024-compute} of completions of 192 FBT scenario variants from \citet{trott2023do}.
We apply vector steering \cite{subramani-etal-2022-extracting, rimsky-etal-2024-steering} to develop a new approach for distinguishing between different drivers of observed FBT performance or failure.
Our findings not only concern the social-cognitive abilities of LLMs but also contribute to long-standing debates in psycholinguistics and developmental psychology about the extent to which these abilities can be learned through linguistic bootstrapping.

\section{Background \& Related work} \label{background}
The concept of ToM, also known as mindreading, is classically defined as the capacity to attribute mental states to others and oneself, to explain and anticipate behaviour \cite{premack1978does}. 
ToM is held to be a precondition for many aspects of social and cultural living, including communication \cite{Sperber2000meta}, forming and maintaining social networks \cite{sutcliffe2007rela}, and cultural learning \cite{tomasello2014ultra}. 
The first tests based on fictional characters with different knowledge states were described by \citet{flavell1968development}, a format that was further developed by \citet{dennett1978beliefs} and \citet{wimmer1983beliefs}, before the most renowned FBT format involving two dolls named Sally and Anne was introduced by \citet{BARONCOHEN198537}.
Variants of this test have been implemented in experiments with infants \cite{barone2019infants}, adults \cite{Meinhardt2011tfb}, non-human animals \cite{Call2008chimptom, vanderVaart2012}, and computer models \cite{arslan2017falsebelief, rabinowitz2018machinetom}. 
There is consensus that the FBT captures an essential part of ToM competence.
Test-retest reliability is reportedly high \cite{hughes2000good} and correlations with real-world social abilities (e.g. detecting lies, understanding non-literal language, \emph{faux pas} avoidance) are well-documented \cite{beaudoin2020systematic, apperly2010mindreaders}. 
However, there is also debate about the actual cognitive requirements for passing the FBT; for an overview and a rebuttal see \citet{jacob2020false}. 
\citet{scott2017early} argue that toddlers and even infants under the age of 2 can pass an FBT once reliance on complex language is removed, whereas others have argued that only language-based tests suffice to assess actual false belief understanding \cite{heyes2014falsebelief}. 

After training on vast amounts of language data, LLMs impress on numerous tasks beyond natural language processing \cite{qin2025largelanguagemodelsmeet}. 
The discovery of emergent abilities on cognitive and behavioural tasks has sparked numerous studies in ``machine psychology'' \cite{hagendorff2024machinepsychology}, targeting, for example, the evolution of language \cite{kouwenhoven-etal-2025-searching, kouwenhoven2025shaping} or strategic \cite{gandhi2023strategic}, emotional \cite{sabour-etal-2024-emobench}, and moral \cite{oh2025moral} reasoning.
Much interest has gone out to the ToM capabilities of LLMs, following roughly three sorts of approaches: building test procedures and benchmarks from text-based paradigms in experimental psychology \cite[e.g,][]{kosinski2024evaluating, wu-etal-2023-hi, chen-etal-2024-tombench}, designing alternative (situated) tests for LLMs \cite{kim-etal-2023-fantom, chan2024negotiationtombenchmarkstresstestingmachine}, and focusing on enhancing performance \cite{moghaddam2023boosting, jin2024mmtomqa}.
Large proprietary models in particular have been shown to perform on par with older children \cite{van-duijn-etal-2023-theory} and adults \cite{strachan2024testing}. 
\citet{trott2023do} have used such a comparison to argue that part of what it takes to pass FBTs can be learned from exposure to large-scale language data, whereas other parts seem to rely on human qualities not possessed by LLMs. 
Concurrent work further strengthens this comparison by showing that, for a range of LLMs, distributional statistics are in principle \textit{sufficient} to develop sensitivity to belief states in FBT reasoning \cite{trott2026languagestatisticsfalsebelief}.
Approaches that apply interpretability techniques have been able to isolate special subnetworks in models that seem to have specialised for FBT reasoning \cite{Wu2025sparseparameters}.

Early implementations of the FBT for end-to-end memory network (MemN2N) \cite{grant2017memoryaugmented, nematzadeh-etal-2018-evaluating} were already criticised for their sensitivity to problems with answer matching and data leakage that allowed superficial heuristics \cite{le-etal-2019-revisiting}. 
These problems persisted in LLM research. 
\citet{ullman2023largelanguagemodelsfail} and \citet{shapira-etal-2024-clever} showed that small modifications in test formulations caused models to fail questions they previously answered correctly, suggesting success was narrow and contingent on surface patterns. 
For a full overview of these issues, see \citet{hu2025re-evaluatingToM}; for further theoretical reflection, see \cite{goldstein2024does}.
Our work resonates with a broader movement calling for careful design and thorough analyses before drawing normative conclusions about the cognitive abilities of LLMs \cite{Frank2023babysteps, Ivanova2025evaluatellms, Mitchell2026evaluating}.

\section{Methods}

\subsection{Tasks}
\label{sec:tasks}
Our analyses build on the FBT battery from \citep{trott2023do}, later incorporated into the EPITOME dataset \cite{jones-etal-2024-comparing-humans}. 
This is a suitable battery since 
\begin{enumerate*}[label=(\arabic*)]
    \item its scenarios are \textit{not} present in the OLMo 2 and OLMo 3 training data, which we rule out through extensive investigation into data leakage (\autoref{app:leakage}), and 
    \item it is counterbalanced to avoid superficial task-solving based on training priors or surface-level heuristics \cite{liu-etal-2024-lost}. 
\end{enumerate*}

The task contains 12 scenarios that conform to the original FBT structure \cite{wimmer1983beliefs}, i.e., a main character places an object at a Start location and a second character moves the object to an End location. 
The experimental question then probes the test subject's (c.q. the LLM's) assessment of where the main character believes the object is. 
Crucially, for each scenario, there are 16 versions that differ in four dimensions (observation condition, knowledge cue, and two location mentions), resulting in 192 test variations: 

\begin{itemize}[leftmargin=*, align=left, font=\bfseries, parsep=0pt, itemsep=1pt, topsep=1pt]
    \item[Knowledge state --] True Belief: the main character is present (and thus sees) when the second character is moving the object. False Belief: the main character is absent (and thus does not see) when the object is being moved.
    \item[Knowledge cue --] Explicit: the propositional attitude of the main character is explicated in the experimental question (X \emph{thinks} the book is in the \dots). Implicit: an action verb is used instead (e.g., X \emph{goes} to get the book from the \dots).
    \item[Location mentions --] Start and End locations are mentioned twice in each scenario, alternating first and most recent mentions (creating two axes). 
\end{itemize}

\begin{figure}[tb]
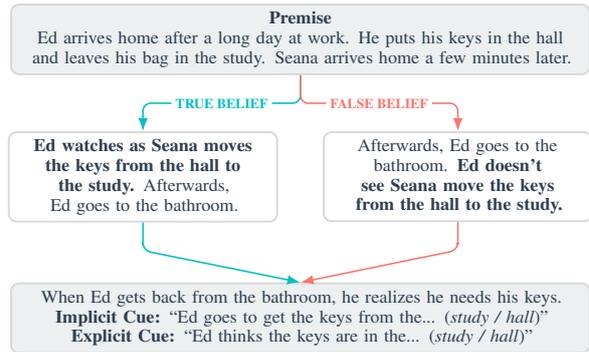

\centering
\includestandalone[width=\columnwidth]{figures/example}
\caption{Example False Belief task from \citet{trott2023do}. A detailed version is visible in \autoref{app:taskmodeldetail}.}
\label{fig:example}
\end{figure}

\noindent An example scenario is shown in \autoref{fig:example}.
\citet{trott2023do} reported that humans correctly answered False Belief questions in $83\%$ of the cases (which aligns with the original finding for the oldest group in \citealp{wimmer1983beliefs}); we include their scores as a human baseline in our plots. 
Our primary manipulation investigates the effect of changing the knowledge state and the knowledge cue. 
Location mentions serve as a control for the potential effect of recency or primacy biases \cite{liu-etal-2024-lost, mina-etal-2025-cognitive}.

\subsection{Computing probabilities} 
We probe an LLM's assessment of the scenario by computing the probability of preset completions. 
Concretely, we compute the conditional probability of the Start and End locations, corresponding to the False and True Belief, given the cue context. 
Since all tested models use Beginning-of-Word tokenisers that may split the target words into variable-length subword units, we must ensure that the computed probability reflects the target word as a complete, whitespace-delimited unit rather than a prefix of a longer word.
We correct for this following \citet{pimentel-meister-2024-compute}, by normalising subword token probabilities with a marginalisation factor (see \autoref{app:probabilities} for details).
The location word with the highest probability acts as the model's prediction.

\subsection{Models}
We test a suite of 17 different models from 6 families at varying levels of openness (see~\autoref{app:taskmodeldetail} for a full list and the detailed computational setup).
Our subsequent analysis is split into two steps. 
First, in \autoref{sec:results}, we assess performance on the FBT across all models to enable LLM-generic analyses.
Here, we distinguish between model variants based on \emph{intended capability}, i.e., base, instruct, and reasoning.
We do so since models differ in how they achieve said variants, using different data and training regimes. 
For example, Llama 3 instruct models were finetuned using \texttt{SFT} and \texttt{DPO}, while K2-V2 instruct only used \texttt{SFT}, and OLMo 2 and 3 underwent \texttt{SFT}, \texttt{DPO}, and \texttt{RLVR}. 
Second, in \autoref{sec:case}, we trace OLMo 2's acquisition of FBT competence across pre- and mid-training, and unravel how this is affected during post-training.

\subsection{Analyses}
Our analyses use Bayesian Regression Models as implemented in the brms package \cite{buckner2021brms} and Bayesian two-sample t-tests as implemented in the BayesFactor package \cite{Bayes2023Morey} in R \cite{r2025}.
We focus on correctly predicted answers (\emph{correct}) for a True Belief or False Belief \emph{knowledge state} that is prompted with a specific \emph{knowledge cue}.

We fit a Bayesian Multilevel Logistic Regression model with a logit function:
\begin{align*}
correct \sim\; & model\_size * variant * \\
& knowledge\_state * knowledge\_cue \\
& + (1 \mid model\_family\_version)
\end{align*}
It predicts whether a given model size (in Billions), training variant (Base, Instruct, Reasoning), knowledge state (False Belief, True Belief), knowledge cue (Explicit, Implicit), and their interactions will correctly answer the question (binary). 
A random intercept for the model family and its version (e.g., llama\_3.1, olmo\_3) is included because these versions may differ in baseline accuracy.
Each model is fitted using 4 chains, each with 4000 iterations and a warm-up of 2000.
Variable coefficients indicate the direction and magnitude of an effect; we consider effects reliable (i.e., significant) when their $95\%$ credible intervals (CI) \emph{exclude} zero, indicating that at least $95\%$ of the posterior probability mass falls on one side of zero.

\section{Results}
\label{sec:results}
Despite substantial variance across model families and versions ($\beta=0.34$), likely reflecting differences in architecture, training data, and optimisation techniques, we still observe LLM-generic effects that we discuss now.

\subsection{Does model size predict ToM reasoning?}
\label{sec:model_size}
It is generally claimed that scaling model size and pre-training data positively correlate with downstream performance and alignment with humans \cite[e.g.,][]{hoffmann2022trainingcomputeoptimallargelanguage, ren-etal-2025-large}.
Here, we address whether these carry over to ToM reasoning as suggested in prior work \cite{kosinski2024evaluating}. 
We argue that to assess first-order ToM, performance across both False and True Belief scenarios should be taken into account, since flexibility to solve the task irrespective of the Knowledge State warrants a much higher degree of generalisability than considering False Belief scores alone.

\begin{figure}
    \centering
    \includegraphics[width=1\linewidth]{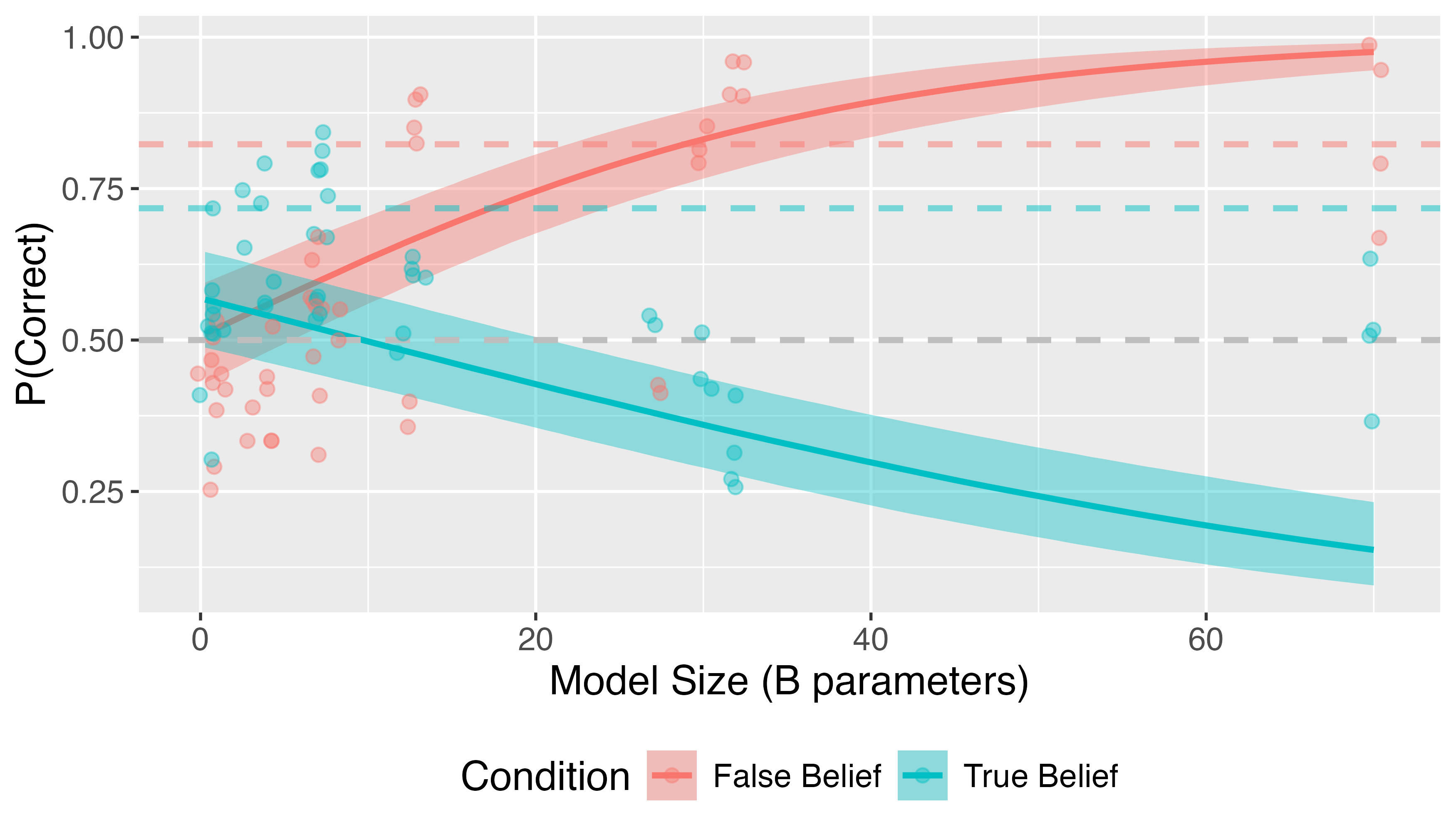}
    \caption{The effect of model size on the probability of being correct given a knowledge condition \emph{without} any other interaction effects. Scaling positively influences False Belief performance but not True Belief. Dashed coloured lines indicate human performance.}
    \label{fig:modelsize-state-performance}
\end{figure}

The Bayesian model confirms that scaling up model size improves the log-odds of correct responses.
Specifically, \emph{without} accounting for any other effects, each additional unit (Billions) of model size increases log-odds performance by 0.05 ($\beta=.05, CI=[.04, .07]$). 
This aligns squarely with findings reported by \citet{trott2026languagestatisticsfalsebelief}.
Interestingly, however, this effect is not uniform across knowledge states, revealing a stark difference between False and True Belief tasks (\autoref{fig:modelsize-state-performance}). 
Scaling strongly benefits False Belief, though it \emph{harms} performance in the True Belief cases. 
While the classical developmental pattern is the reverse (false beliefs being harder than true beliefs in young children; \citealt{wellman2001meta}), like models, adults appear to find the True Belief scenarios somewhat more difficult than the False Belief ones \cite{trott2023do}.
This asymmetry is far more pronounced in models than in humans. 
Although \citet{trott2026languagestatisticsfalsebelief} show that larger models are better predictors of human responses, this asymmetry may explain why even these larger models cannot fully account for the human data.
Returning to the question of whether model size predicts ToM reasoning, we observe that increasing model size does not have a strictly positive effect when both True Belief and False Belief scenarios are taken into account.

\subsection{Explicating propositional attitudes}
\label{sec:propositional}
Earlier work reported that for GPT-3, FBT performance improved when propositional attitudes were explicated (e.g., ``John thinks the wine is in the\dots'') as opposed to when they were left implicit \cite[e.g., ``John goes to get the wine from the\dots'';][]{trott2023do}.
Our results corroborate this effect, expand it to a broader range of models, and further specify its occurrence.
Across all model sizes, implicit cues are consistently harder than explicit cues: \emph{without} taking other effects into account, the log-odds of a correct response decrease by 1.11 units for implicit versus explicit cues ($\beta=-1.11, CI=[-1.40, -0.84]$).
In our sample, this effect is equally strong across model sizes ($\beta=-0.01, CI=[-0.03, 0.00]$). 
However, \citet{trott2026languagestatisticsfalsebelief} use \textit{additional} models and find that this effect even grows with parameter count.

\begin{figure}
    \centering
    \includegraphics[width=1\linewidth]{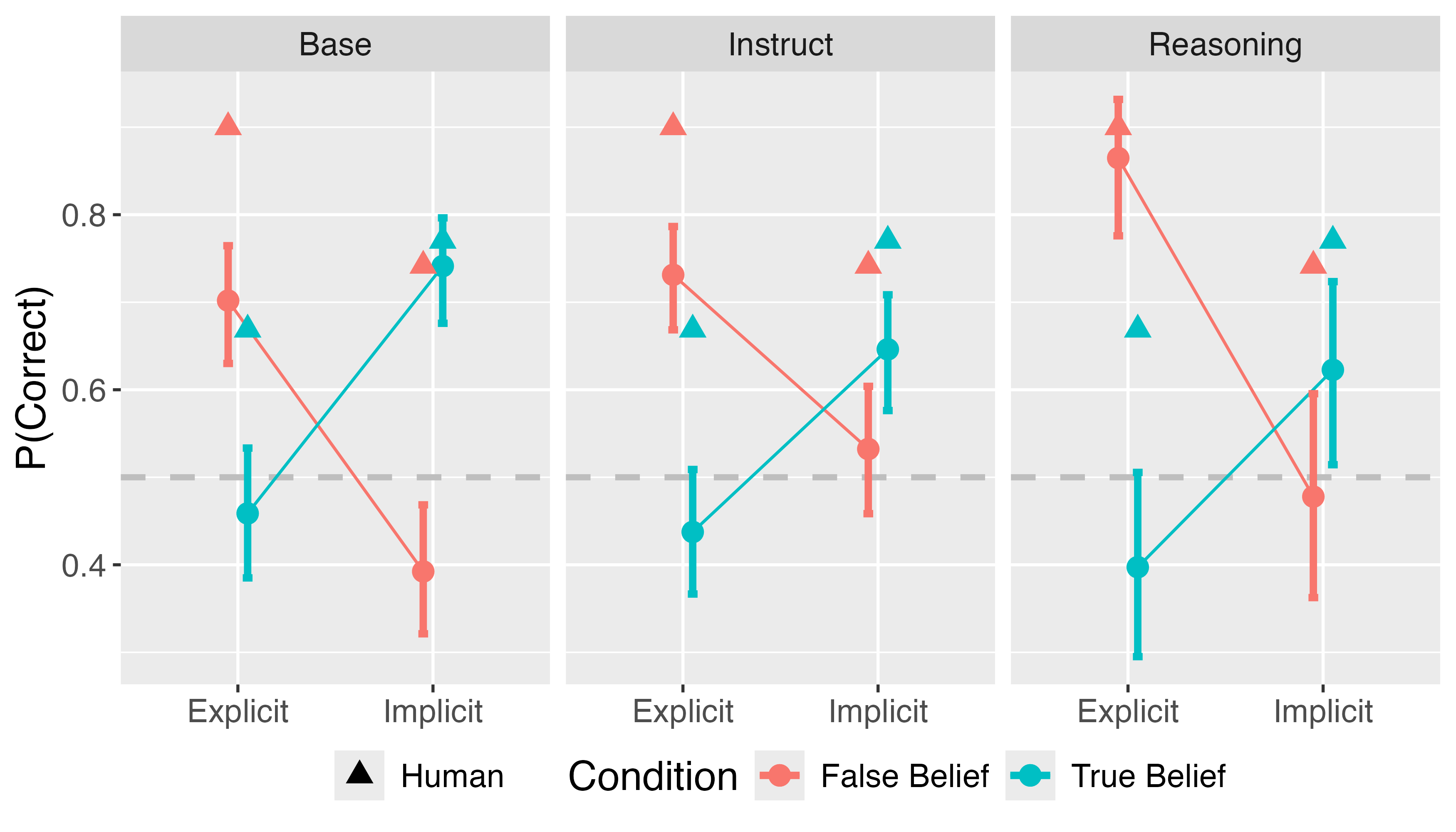}
    \caption{The probability of predicting the location correctly for model variant, knowledge state, and cue. Triangles indicate the average human performance.}
    \label{fig:interaction-variants}
\end{figure}

Informed by the effects of scaling and implicit cues, we now aim to unravel why these cause the stark difference between True and False Belief scenarios. 
The answer lies in a striking crossover interaction between knowledge state and cue ($\beta=2.07, CI=[1.70, 2.46]$, \autoref{fig:interaction-variants}), meaning that implicit cues impair False Belief performance but facilitate True Belief performance. 
Vice versa, explicit cues facilitate False Belief but hurt True Belief performance.
This suggests that explicating propositional attitudes (X \emph{thinks}) fundamentally alters response patterns and may explain why solving TB fails to benefit from scaling. 
Since the performance gap between FB and TB grows with model size, it is likely driven by a pattern learned from the data that is amplified as the model scales.
Though it is tempting to attribute this solely to surface-level heuristics or pattern-matching \cite[e.g.,][]{ullman2023largelanguagemodelsfail, shapira-etal-2024-clever}, our results suggest a more nuanced mechanism.
The crossover pattern indicates that models are sensitive to two interacting factors: (1) learned associations with stereotypical False Belief scenarios, and (2) the presence of explicit propositional attitudes themselves. 

Critically, this does not necessarily imply an absence of genuine ToM reasoning. 
Rather, it suggests that explicit propositional attitudes and learned scenario patterns interact with such reasoning in ways that can either facilitate or interfere with the process of arriving at a correct prediction. 
Interestingly, \citet{trott2026languagestatisticsfalsebelief} report that the use of non-factive verbs affects human FBT performance \textit{in similar ways} to those of LLMs. 
They ascribe this effect to \textit{anti-presupposition}, i.e., the idea that `Alex thinks X' weakly implies `NOT X' \cite{chemla2008anti-presup}.
This is in line with \citet{white2018prop}'s analysis of \emph{think} as a representational non-factive propositional attitude verb.
Compare the following two sentences:
\begin{flushleft}
     \hspace{1.5em} (a) Alex thinks that Bo went to the store.\\
     \hspace{1.5em} (b) Alex knows that Bo went to the store.
\end{flushleft}
Sentence (a) is pragmatically fitting when the speaker does not know whether Bo went to the store or not, or knows for certain that Bo did not go. 
Yet it does \emph{not} fit a context where the speaker knows for certain that Bo went to the store; here (b) would be more appropriate \cite[c.f. also][]{stalnaker1973presuppositions}. 
Because of this distinction, \emph{think} can be expected to be frequently used to mark epistemic divergence rather than congruency within communicative settings. 
Applied to LLMs and the narrow context of the FBT, when \emph{think} appears, it may activate a learned pattern of opposing what is actually known to be the case. 
The presence of the non-factive propositional attitude verb \emph{think} may thus bias the LLM towards predicting a \textit{contrast} between the agent's belief and the object's location (`NOT X').
This benefits test performance in the False Belief condition but impairs it in True Belief scenarios.
We further investigate this account in \autoref{sec:case}.

\subsection{Does post-training act as a cooperative pressure?}
\label{sec:post-training}
Existing work draws a parallel between post-training and the cooperative pressures that shape human communication \cite{van-duijn-etal-2023-theory}. 
For humans, communication is fundamentally cooperative and relies on the ability and willingness to engage in mental coordination \cite[e.g.][]{grice1975logic, verhagen2025grammar}.
This willingness to engage in cooperation to achieve successful communicative interactions is constantly rewarded \cite{tomasello2010origins}, while failing to do so results in punishment in the form of social exclusion \cite{davidbarrett2016language}.
Post-training essentially induces analogous cooperative principles, rewarding helpful responses and penalising failures to coordinate with the user's intent.
\citet{van-duijn-etal-2023-theory} therefore hypothesised that instruction-tuned LLMs, like humans, bank on the capacity to coordinate with an interaction partner's perspective, benefiting ToM tasks that depend on precisely this capacity. 

Here, we empirically test whether this parallel holds by examining the effects of different model variants. 
If the parallel exists, we hypothesise that post-trained model variants that follow instructions will show this most notably, as they are rewarded for interactive behaviour. 
While the analogy is less clear for models that have been post-trained for reasoning, one might expect these models to attend more carefully to prompt details (``\dots think step-by-step \dots'').
However, recent research has shown revealed that reasoning remains fragile to surface-level input variations \citep{mirzadeh2025gsmsymbolic}, suggesting that reasoning may also amplify heuristic (i.a., pattern matching, pick the last location) responses.


Since post-training is done sequentially (i.e., first pre-training \texttt{Base}, then \texttt{SFT}, then \texttt{DPO}, etc.), we use checkpoints that reflect this structure (i.e., Base, Instruct, and Reasoning). 
Importantly, the effects reported should be interpreted as the cumulative effect of additional stages after pre-training, \emph{not} the incremental effect of each training stage individually. 
Moreover, reported values represent performance at the mean model size, averaging over models in that condition and variant. 

Given the interaction between knowledge states and propositional attitudes, we now examine how this differs after post-training.
We find that post-training affects FBT accuracy.
Both Instruct and Reasoning variants improve relative to Base models (Instruct: $\beta=0.33$, CI$=[0.07, 0.57]$; Reasoning: $\beta=0.74$, CI$=[0.10, 1.39]$), aligning with work showing that post-trained models (i.a., OLMo 2) are among the most human-aligned \cite{studdiford2025uncoveringcomputationalingredientshumanlike}.
However, these overall gains mask important differences in how training variants interact with task characteristics.
\autoref{fig:interaction-variants} also reveals that the crossover pattern discussed above is not uniform across training variants. 
In Base models, it is most pronounced as implicit cues strongly impair False Belief while substantially facilitating True Belief. 
Instruct training partially rescues the implicit False Belief deficit ($\beta=0.37, CI=[0.02, 0.72]$), attenuating the crossover. 
Training to induce Reasoning, by contrast, worsens the implicit False Belief impairment ($\beta=-1.09, CI=[-1.93, -0.24]$), while also reducing True Belief performance ($\beta=-0.93, CI=[-1.73, -0.11]$), amplifying sensitivity to knowledge cue type dramatically.
This suggests that reasoning training specialises models for explicit False Belief reasoning at the cost of broader ToM robustness.

Thus, the answer to whether post-training acts as a cooperative mechanism \citep{van-duijn-etal-2023-theory} is nuanced: while improving FBT performance overall, the gains are modulated by stronger interaction effects, and only robust performance across conditions would indicate true ToM reasoning abilities.

\section{Case study: Traces of social intelligence in OLMo 2}
\label{sec:case}
While previous analyses focused on a large sample of models, we now narrow the analyses down to OLMo 2 models only. This allows us to 
\begin{enumerate*}[label=(\arabic*)]
    \item limit data contamination risks, as we are certain that the stimuli are \textit{not} present in any of the training data (\autoref{app:leakage}), and
    \item deepen our investigation into the influence of fine-tuning techniques (e.g., \texttt{SFT}, \texttt{DPO}, \texttt{RLVR}) on the observed cross-over effect, since we know exactly how OLMo is trained.
\end{enumerate*}

\subsection{Pre-training dynamics} 
We follow \citet{mahowald2024dissociating} and ask whether acquiring low-level formal language competencies naturally extends to functional linguistic capabilities.
In line with previous analyses revealing early and sudden syntax acquisition \cite{chen2024sudden}, OLMo 2 acquires formal linguistic capabilities, as measured by achieving $>80\%$ accuracy on the BLIMP benchmark \cite{warstadt2020blimp}, after observing only $0.05\%$ of the total number of training tokens. 
In contrast, it takes about 25 times longer ($12.5\%$ of the data observed) for the model to achieve above-random performance on the False Belief task. 
Moreover, BLIMP performance remains stable throughout training, whereas ToM performance fluctuates, further hinting at distinct underlying drivers \cite[][more details in Appendix~\ref{app:linguistic-capability}]{hanna2026formal}. 
Though language acquisition may scaffold ToM in children \cite{milligan2007language, de2014role}, we cannot conclude that merely exposing LLMs to additional tokens will automatically lead to the emergence of ToM. 
However, our findings suggest that formal linguistic capabilities may be required for functional understanding and that post-training objectives can further elicit them.
This aligns with recent evidence that LLM-brain alignment tracks formal competence earlier and more strongly than functional competence \cite{alkhamissi-etal-2025-language}.

This leaves us with the question of when OLMo 2's FBT capability will evolve, as investigated through extensive analysis of its learning dynamics.
The training of OLMo 2 consists of a pre-training phase, a mid-training phase that uses model merging, and a fine-tuning phase using \texttt{SFT}, \texttt{DPO}, and \texttt{RLVR} \cite{olmo20252olmo2furious}. 
For the pre-training stage of each model size, we sample 50 checkpoints evenly spaced by token count, after confirming that our sampled checkpoints yield a representative learning curve of an exhaustive sweep over all checkpoints for the 7B model (\autoref{app:traces}). 
Similarly, we established that mid-training, intended to embed models with specialised knowledge through model merging, does not yield meaningful changes, and we omit this stage from further analyses.

\begin{figure}[t]
    \centering
    \includegraphics[width=\linewidth]{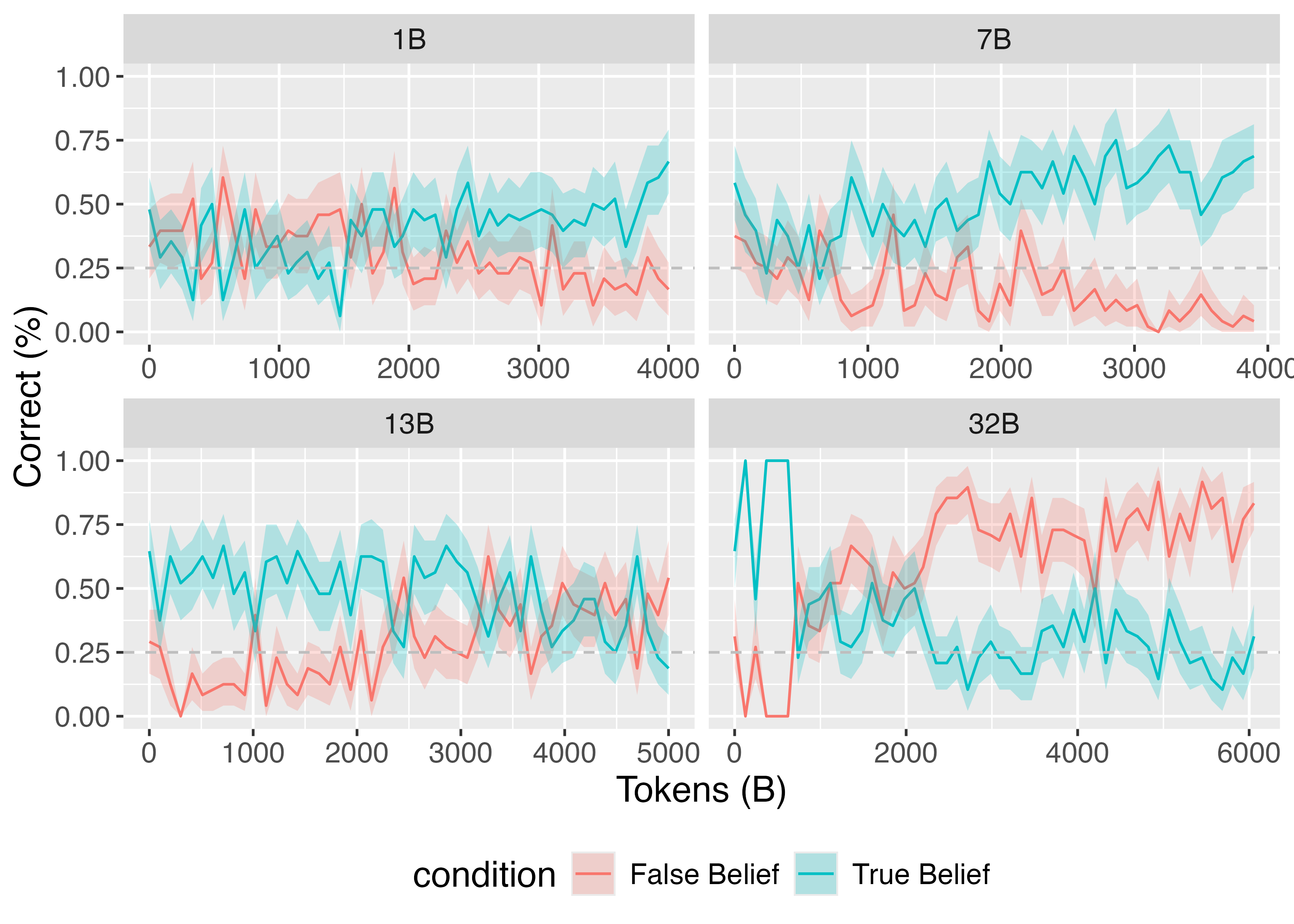}
    \caption{Strict performance during pre-training for different model sizes. Shaded areas indicate $95\%$ CI.}
    \label{fig:strict_traces}
\end{figure}

Moving beyond the effect of knowledge cues, we now aggregate across them to assess robustness to knowledge state. 
This means a response is considered correct only if both the implicit and explicit cue conditions are answered correctly for a given scenario and location (``strictly'' correct).
\autoref{fig:strict_traces} reveals the learning traces for all OLMo 2 model sizes during pre-training. 
Consistently, True and False Belief performance differ significantly and show a cross-pattern that evolves during training.
For small models, False Belief scenarios are more difficult than True Belief scenarios, while larger models show the opposite effect.
This indicates that models are sensitive to the main character's knowledge state but in different respects.  
Following the call for robust human-centred evaluation \cite{Ivanova2025evaluatellms, Mitchell2026evaluating}, we argue once again that attributing ToM to LLMs is fair only if the knowledge state has no impact on the evaluation. 

\begin{table}[t]
\centering
\begin{tabular}{llllll}
                                &       & \textbf{1B}   & \textbf{7B}   & \textbf{13B}  & \textbf{32B}  \\ 
  \toprule
    \multirow{2}{*}{\textbf{FB}}& Imp   & .250          & .042          & .542          & .833          \\
                                & Exp   & .292          & .562**        & .938**        & .979*         \\
  \midrule
    \multirow{2}{*}{\textbf{TB}}& Imp   & .771          & 1.00          & .896          & .750          \\
                                & Exp   & .750          & .688**        & .188**        & .354**        \\
   \bottomrule
\end{tabular}
\caption{Mean accuracy at final pre-training checkpoint. Significance markers indicate a moderate (*$BF_{10}>3$) or strong (**$BF_{10}>30$) difference between paired Explicit (Exp) and Implicit (Imp) questions.}
\label{tab:bayesian_results}
\end{table}

To unravel why we observe such stark differences in performance across different knowledge states, we analyse them by knowledge cue. 
Bayesian T-tests testing whether the mean accuracy differs across knowledge cues at the last checkpoint for each model confirm our earlier finding (\autoref{sec:propositional}) that explicating propositional attitudes hurts False Belief reasoning, whilst implicit reasoning improves True Belief tasks (\autoref{tab:bayesian_results}). 
This shows tentative similarities to the finding that LLM representations of explicit prompts better correlate with brain activations than those of nonsensical, noisy prompts \citep{ren-etal-2025-large}.
Importantly, we want to stress that the collapse is not due to an apparent primacy or recency bias of OLMo 2 since location mentions are balanced. 
Our observation is largely consistent across model sizes and again confirms that scale helps with False Belief but hurts True Belief, except for the 1B model. 
Yet, given its stable low performance on the False Belief task, it may simply not be up to the task at all and thus not affected by changing the knowledge cue. 

\begin{figure*}[ht]
    \centering
    \includegraphics[width=1\linewidth]{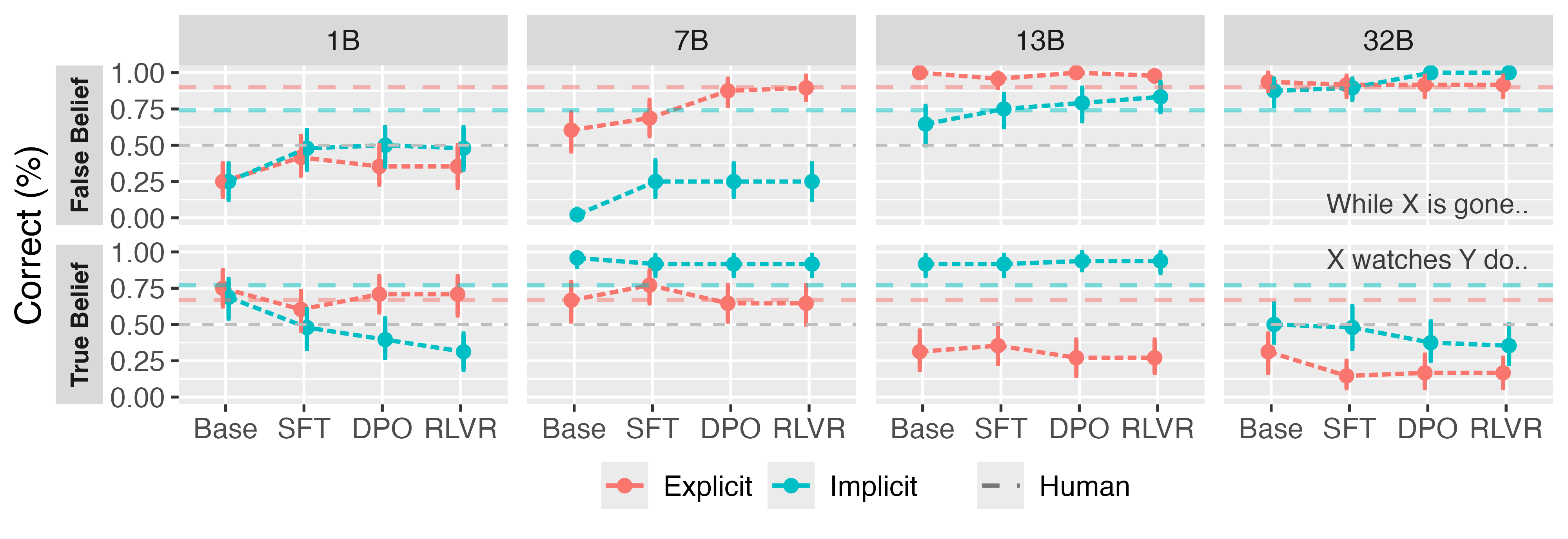}
    \caption{The percentage of correct answers in different base or post-training phases for differently sized OLMo 2 models. Coloured dashed lines indicate human performance, and the bars indicate 95\% confidence intervals.}
    \label{fig:fb_olmo_performance}
\end{figure*}

Returning to the question of how OLMo 2's ToM ability evolves, we observe that strict False and True Belief reasoning diverge during pre-training.
As such, pre-training leads to the earlier-observed interaction between knowledge states and cues, driven by OLMo 2's ToM ability, which is not robust to adversarial cases involving explicit True Belief or implicit False Belief. 
While OLMo 2 does not encounter the exact test scenarios during pre-training, our scenarios follow a format very similar to the canonical Sally-Anne format \cite{BARONCOHEN198537}. 
As argued in \autoref{sec:propositional}, explicating propositional attitude verbs also drives model responses.
The combination of both implies that OLMo 2 may acquire stereotypical response patterns tied to mental-state vocabulary that can outweigh other scenario semantics. 
Pre-training language models on carefully curated datasets, excluding canonical FBT formats, would be required to further crystallise this hypothesis. 

\subsection{Post-training dynamics}
\label{sec:post-training-mechanism}
Earlier LLM-generic analyses showed that post-training has a nuanced effect on the cross-over pattern between the knowledge state and the implicit/explicit cues (\autoref{sec:post-training}). 
The open-source nature of OLMo 2 enables us to investigate which post-training methods drive the cross-over effect.  
 
Again, we observe a dichotomy between True and False Belief accuracy across model sizes (\autoref{fig:fb_olmo_performance}). 
For False Belief tasks, smaller models benefit somewhat from post-training, especially using \texttt{SFT} and \texttt{DPO}, where we observe the smallest effect on the largest model.
Similar findings have been described in a similarity judgement task \citep{studdiford2025uncoveringcomputationalingredientshumanlike} and in investigations towards generating diverse stories \cite{peeperkorn2025mindgapconformativedecoding}.
For True Belief tasks, however, post-training has little effect, with two notable exceptions. 
First, post-training has a pronounced negative effect on the True Belief performance of the 32B model.
Even though our exact task is not present in the training data, this model may be so large that it has learned the general FBT structure and is doing pattern matching. 
Using the pattern as a heuristic could also explain why post-training does not add, and sometimes hurts; if the heuristic is too strong, additional, deviating signals are ignored \citep{ullman2023largelanguagemodelsfail, shapira-etal-2024-clever}.  
Second, the 1B model under the implicit cue also shows a considerable drop in True Belief accuracy after post-training.
Since this model has not exactly acquired what it takes to solve the FBT (i.e., it performs below chance for False Belief scenarios), further `pushing' it with additional cooperative signals likely only leads it to form misconceptions about knowledge states.
Thus, for OLMo 2, False Belief Scenarios seem to benefit from \texttt{SFT} and \texttt{DPO} post-training. 
While for True Belief scenarios, post-training hurts or has no effect. 


\section{Steering with \emph{think} vectors} \label{sec:steering}
\begin{figure}[t]
    \centering
    \includegraphics[width=\linewidth]{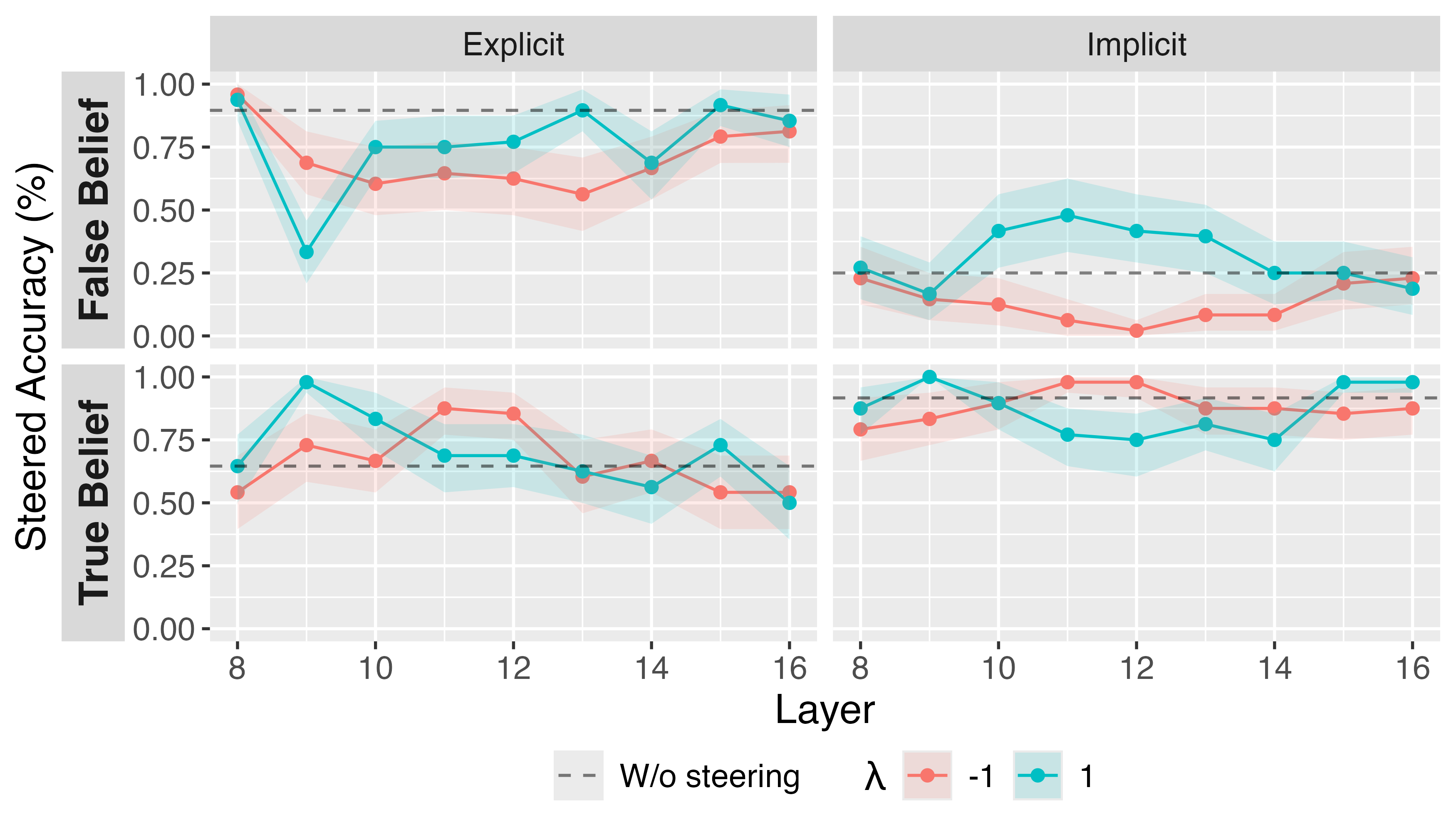}
    \caption{Model steering results. Dotted lines indicate \emph{unsteered} accuracy and coloured lines indicate positively (green) or negatively (red) \emph{steered} performance.}
    \label{fig:steering}
\end{figure}

Motivated by the strong influence of knowledge cues, we apply model steering \cite{subramani-etal-2022-extracting, rimsky-etal-2024-steering, siddique2025shifting} to test whether OLMo 2 7B Instruct, selected for its stark explicit–implicit gap (\autoref{fig:fb_olmo_performance}), encodes a representational direction that can causally influence predictions.
Model steering (intuitively) captures a model's contrast between two sets of opposing prompts by embedding them and computing their difference. 
The resulting vector can be used to steer future predictions. 
Specifically, we extract steering vectors by computing the difference in last token hidden-state activations between paired explicit and implicit scenarios at layers 8–16, as this region proved most susceptible to steering in preliminary tests and in previous work \cite{subramani-etal-2022-extracting}.
Since pairs differ only in whether or not propositional attitudes are explicated, these vectors isolate the representational contrast, or direction, between knowledge states––hence \emph{think} vector. 
We inject these scaled ($\lambda$) vectors during inference at intermediate layers: $\lambda=1$ adds the think vector to scenarios, supposedly pushing the model toward explicit-like behaviour, while $\lambda = -1$ subtracts it. 
To prevent information leakage, we use a leave-one-out procedure to ensure no scenario information is encoded in the think vector.
This means that, for each scenario, we construct a think vector using the remaining 11 scenarios. 
Steered accuracy is compared against the unsteered baseline in \autoref{fig:steering}.

Model steering appears most effective at layers 10 to 12, to which we confine our discussion here.
Adding `explicitness' to \emph{implicit} False Belief increases accuracy, while subtracting it decreases accuracy.
Subtracting `explicitness' from \emph{explicit} False Belief decreases performance.
Addition also hurts, but since base performance is at the ceiling and these scenarios are already explicit, it introduces redundancy and may therefore be disruptive.
Our analyses showed that, for explicit True Belief cases, performance was lower (\autoref{sec:post-training-mechanism}). 
In line with those findings, adding a think vector hurts for implicit True Belief cases, while subtracting it here boosts performance.
Adding the think vector in explicit True Belief cases has little effect, whereas subtracting it leads to higher performance, again in line with expectations.
These effects in the expected directions corroborate our suggestion that answering patterns can be rather strongly influenced by information encoded in the propositional attitude verb \emph{think}, most likely contingent on its use in contexts of epistemic divergence in the training data.

\section{Conclusion}
Given ToM's central role in social interaction, we systematically tested the performance of 17 LLMs on the False Belief Test.
Our analyses amount to the main finding that FBT performance of LLMs in our sample shows nuanced associations with scale and post-training, but is not robust to False vs. True and explicit vs. implicit variations. 
It appears that the interaction between learned scenario patterns and information contained in the verb \emph{think} interferes with genuine social reasoning, sometimes in productive and sometimes in detrimental ways.
Our findings inform discussions of the nature of social intelligence in LLMs and humans, and highlight the need for careful evaluation on high-quality tests.

\section*{Limitations}
We highlight three noteworthy limitations of our work.
First, we focus solely on ToM, while the ability to attribute mental states to others is clearly only a subset of socio-cognitive competencies.
Social cognition involves a suite of mechanisms and strategies, ranging from low-level, implicit processes that rely on situated and embodied forms of cognition to reliance on norms, conventions, and common ground. 
All of these mechanisms amount to social competence and are important in everyday human-LLM interactions.
Furthermore, the FBT is only a limited subset of ToM that, in our case, involves first-order mindreading.  
Beyond investigating higher-order cases, to speak of social competence in LLMs, it is pertinent that models perform robustly across a range of tests \cite[e.g., recursive mindreading, strange stories, imposing memory;][]{jones-etal-2024-comparing-humans}.

Second, our finding that explicating propositional attitudes interacts with knowledge states currently relies on two verbs.
These analyses should be extended and corroborated by adding different propositional attitude verbs (e.g., \textit{believe}, \textit{assume}) as test variants.
Circuit analyses as in \citet{Wu2025sparseparameters} may further reveal whether propositional attitude verbs are encoded differently from non-propositional attitudes.

Lastly, recent work has revealed that LLMs may recognise when vector steering is applied \cite{fornasiere2026languagemodelsrecognizedropout}, attributed to introspection by some \cite{lindsey2026emergentintrospectiveawarenesslarge, pearsonvogel2026latentintrospectionmodelsdetect}.
As of yet, it is unclear whether and how this alters LLM inference.
To rule out whether our think vectors influence predictions, further stress testing with attention checks or evaluating factual questions in our scenarios can be conducted in additional analyses.
We leave these analyses to future work.

\bibliography{custom}

\appendix

\section{Computing probabilities}
\label{app:probabilities}
Since all tested models use Beginning-of-Word (BOW) tokenisers that might split the target words into variable-length subword units, we need to ensure that the computed probability reflects the target word as a complete, whitespace-delimited unit rather than a prefix of a longer word.
We correct for this following \citet{pimentel-meister-2024-compute}, as shown in Eqn~\ref{eq:word-prob}.

\begin{align}
    \label{eq:word-prob}
    & p( w \mid \mathbf{w}_{<t} ) = \nonumber \\
    & \quad p( \mathbf{s}^w \mid \mathbf{s}^{\mathbf{w}^{<t}} ) 
    \frac{ \displaystyle \sum_{ \{ s \in \overline{\mathcal{S}}_{\text{bow}} \} } p( s \mid \mathbf{s}^{\mathbf{w}^{<t}} \circ \mathbf{s}^w ) }{ \displaystyle \sum_{ \{ s \in \overline{\mathcal{S}}_{\text{bow}} \} } p( s \mid \mathbf{s}^{\mathbf{w}^{<t}} ) } 
\end{align}

\noindent We compute the probability of word $w$ being generated following its preceding context $\mathbf{w}_{<t}$ by multiplying the probability of generating its constituent subword tokens $\mathbf{s}^w$ conditioned on the subword tokens for the previous words $\mathbf{s}^{\mathbf{w}^{<t}}$, with a marginalisation factor for all possible continuations of the target location word. 
This ensures that we compute the probability of the target location word being delimited by white space rather than continued by additional subwords, which would yield a different word than $w$. 
The marginalisation factor, introduced by \citet{pimentel-meister-2024-compute}, is computed as the ratio between the total probability of tokens that begin a new word $s \in \overline{\mathcal{S}}_{\text{bow}}$ following the predicted token appended to its preceding context $\mathbf{s}^{\mathbf{w}^{<t}} \circ \mathbf{s}^w$, divided over the total probability of all possible word-starting tokens at that position.
This normalisation ensures that we compute a contextual sequence probability, where individual token probabilities are normalised to account for BOW tokenisers.

\section{Detailed False Belief task and tested Models}
\label{app:taskmodeldetail}
The open weight and open source models used in this work are listed in \autoref{tab:model_overview}.

\begin{table*}[ht]
    \centering
    \small 
    
    \begin{tabular}{@{} 
        >{\raggedright\arraybackslash}p{0.18\textwidth}
        >{\raggedright\arraybackslash}p{0.18\textwidth}
        >{\raggedright\arraybackslash}p{0.27\textwidth} 
        >{\raggedright\arraybackslash}p{0.27\textwidth} 
    @{}}
        \toprule
        \textbf{Model}          & \textbf{Authors}              & \textbf{Sizes}            & \textbf{Variants}             \\ 
        \midrule
        OLMo 2                  & \citet{olmo20252olmo2furious} & 1B, 7B (exh), 13B, 32B    & Ckpts, Base, SFT, DPO, RLVR    \\
        \addlinespace
        OLMo 3                  & \citet{olmo2025olmo3}         & 7B, 32B                   & Base, SFT, DPO, RLVR, Thinking \\
        \addlinespace
        Gemma 3                 & \citet{gemma3report}          & 270M, 1B, 4B, 12B, 27B    & Base, IT                        \\
        \addlinespace
        Qwen 3                  & \citet{yang2025qwen3}         & 4B, 30B                   & Base, IT, Thinking              \\
        \addlinespace
        K2-V2                   & \citet{k2team2026k2v2360}     & 70B                       & Base, IT                        \\
        \addlinespace
        Llama 3.1, Llama 3.2    & \citet{grattafiori2024llama3} & 1B, 3B, 8B, 70B           & Base, IT                        \\
        \bottomrule
    \end{tabular}
    \caption{Overview of the models used in our experiments. An exhaustive (exh) checkpoint sweep was done for OLMo 2 7B.}
    \label{tab:model_overview}
\end{table*}

A more detailed overview of the task dimensions and example as explained in \autoref{sec:tasks} can be found in \autoref{tab:task-overview}.

\begin{table*}[ht]
\centering
\small
\begin{tabular}{@{} 
    >{\raggedright\arraybackslash}p{0.15\textwidth}
    >{\raggedright\arraybackslash}p{0.30\textwidth}
    >{\raggedright\arraybackslash}p{0.49\textwidth} 
@{}}
\toprule
\textbf{Dimension} & \textbf{Level} & \textbf{Example} \\
\midrule
Premise & & Ed arrives home after a long day at work. He puts his keys in the hall and leaves his bag in the study. Seana arrives home a few minutes later. \\
\midrule
\multirow{2}{=}[-1.5em]{Knowledge state} 
    & True: character can see that the object is being moved 
    & Ed watches as Seana moves the keys from the hall to the study. Afterwards, Ed goes to the bathroom. \\
    \cmidrule(lr){2-3}
    & False: character cannot see that the object is being moved 
    & Afterwards, Ed goes to the bathroom. Ed doesn't see Seana move the keys from the hall to the study. \\
\midrule
\multirow{4}{=}[-0.5em]{Location mentions} 
    & Start location mentioned first 
    & \ldots puts his keys in the hall and leaves his bag in the study\\
    & Start location mentioned last 
    & \ldots moves the keys to the study from the hall\\
    \cmidrule(lr){2-3}
    & End location mentioned first 
    & \ldots leaves his bag in the study and puts his keys in the hall\\
    & End location mentioned last 
    & \ldots moves the keys from the hall to the study\\
\midrule
\multirow{2}{=}[-0.5em]{Knowledge cue} 
    & Implicit: action verb ``go to''
    & Ed \textbf{goes to} get the keys from the [mask] \\
    \cmidrule(lr){2-3}
    & Explicit: propositional attitude verb ``think'' 
    & Ed \textbf{thinks} the keys are in the [mask] \\
\bottomrule
\end{tabular}
\caption{Overview of task variables and their realisations. Each scenario is constructed by combining one level from each variable, yielding 192 unique items.}
\label{tab:task-overview}
\end{table*}

\paragraph{Computational setup}
We run our models on heterogeneous GPU-enabled hardware, including H100 and A100 GPUs. The total running time for the FB dataset with the largest models is 15 minutes per model and experiment condition. 
Models up to 16B are run in full-precision mode (32-bit floats), larger model sizes are run in 16-bit precision mode.

\section{Traces}
\label{app:traces}
\subsection{Detailed pre-training \& mid-training traces}
\begin{figure*}
    \includegraphics[width=\textwidth]{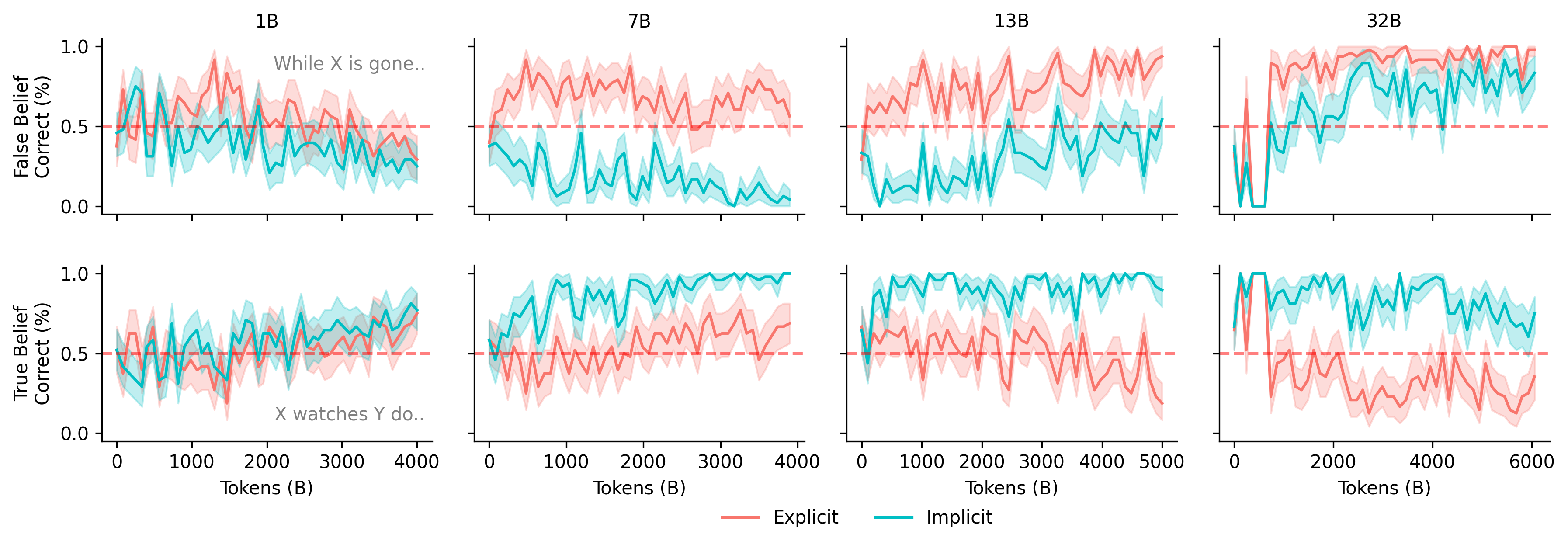}
    \caption{OLMo 2 traces of social intelligence for stage 1. Shaded areas are 95\% confidence intervals.}
    \label{app:fig:traces}
\end{figure*}

\begin{figure*}
    \includegraphics[width=\textwidth]{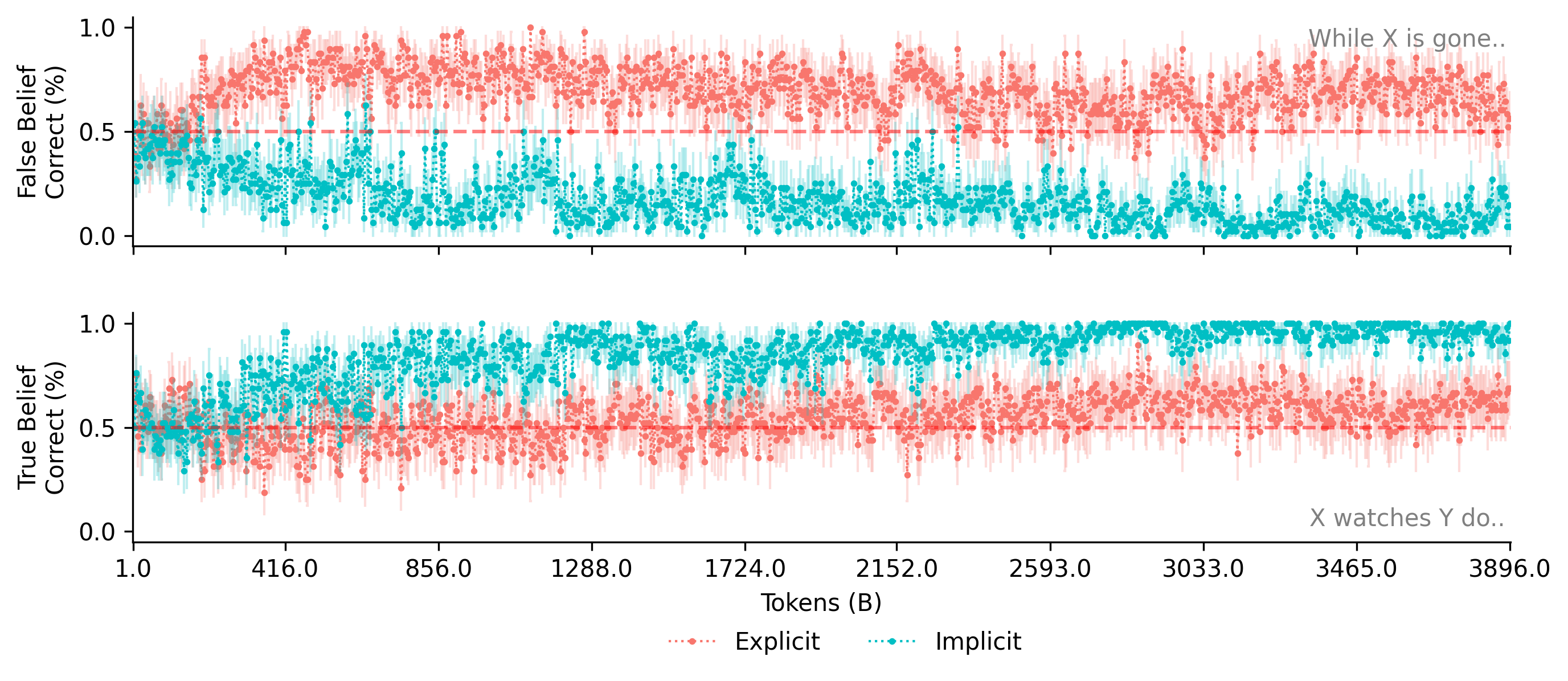}
    \caption{The exhaustive evaluation for stage 1 of OLMo 2 7B checkpoints to trace the emergence of social intelligence.}
    \label{app:fig:7B-traces}
\end{figure*}

\begin{figure}
    \includegraphics[width=\columnwidth]{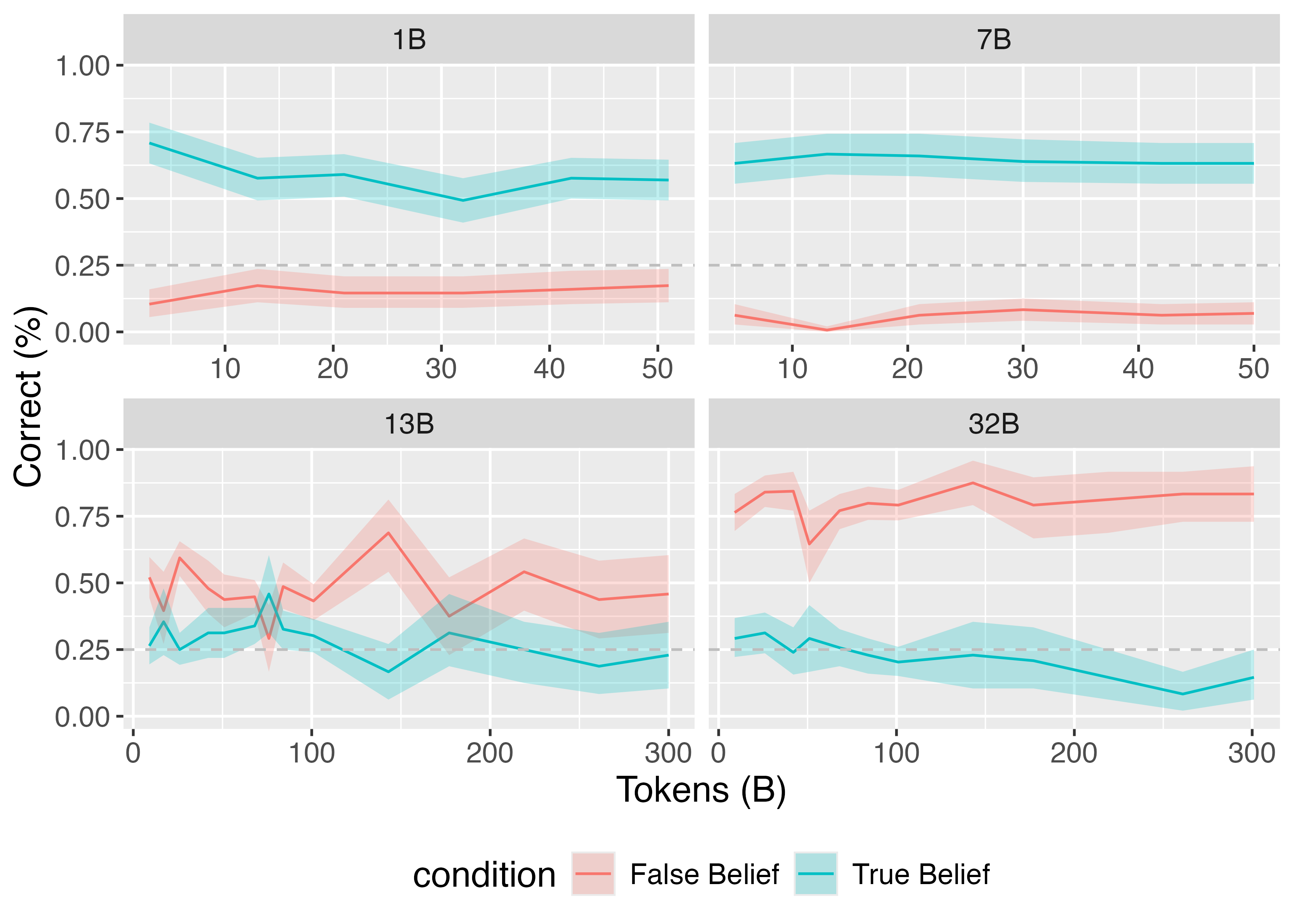}
    \caption{OLMo 2 mid-training traces of strict performance on our task.}
    \label{app:fig:stage2-7B-traces}
\end{figure}

This appendix presents detailed learning curves for OLMo 2. 
We provide a subsampled overview (i.e., we sample 50 checkpoints evenly spaced by token count) of pre-training across all model sizes in \autoref{app:fig:traces} and a complete view (all checkpoints) for the 7B model in \autoref{app:fig:7B-traces}.
The detailed run using all pre-training checkpoints shows considerable variation across checkpoints.
However, we observe the same general trend when subsampling checkpoints: over the course of training, FB performance tends to increase or decrease. 
Hence, we use the subsampled results in the main paper to draw our conclusions about the learning dynamics.
\autoref{app:fig:stage2-7B-traces} shows the strict performance of mid-training. 
There seems to be no clear difference between the start and end of mid-training.
We therefore did not incorporate mid-training analyses into the main body of this paper.

\subsection{Developing linguistic capability}
\label{app:linguistic-capability}
\begin{figure}[t]
    \centering
    \includegraphics[width=\columnwidth]{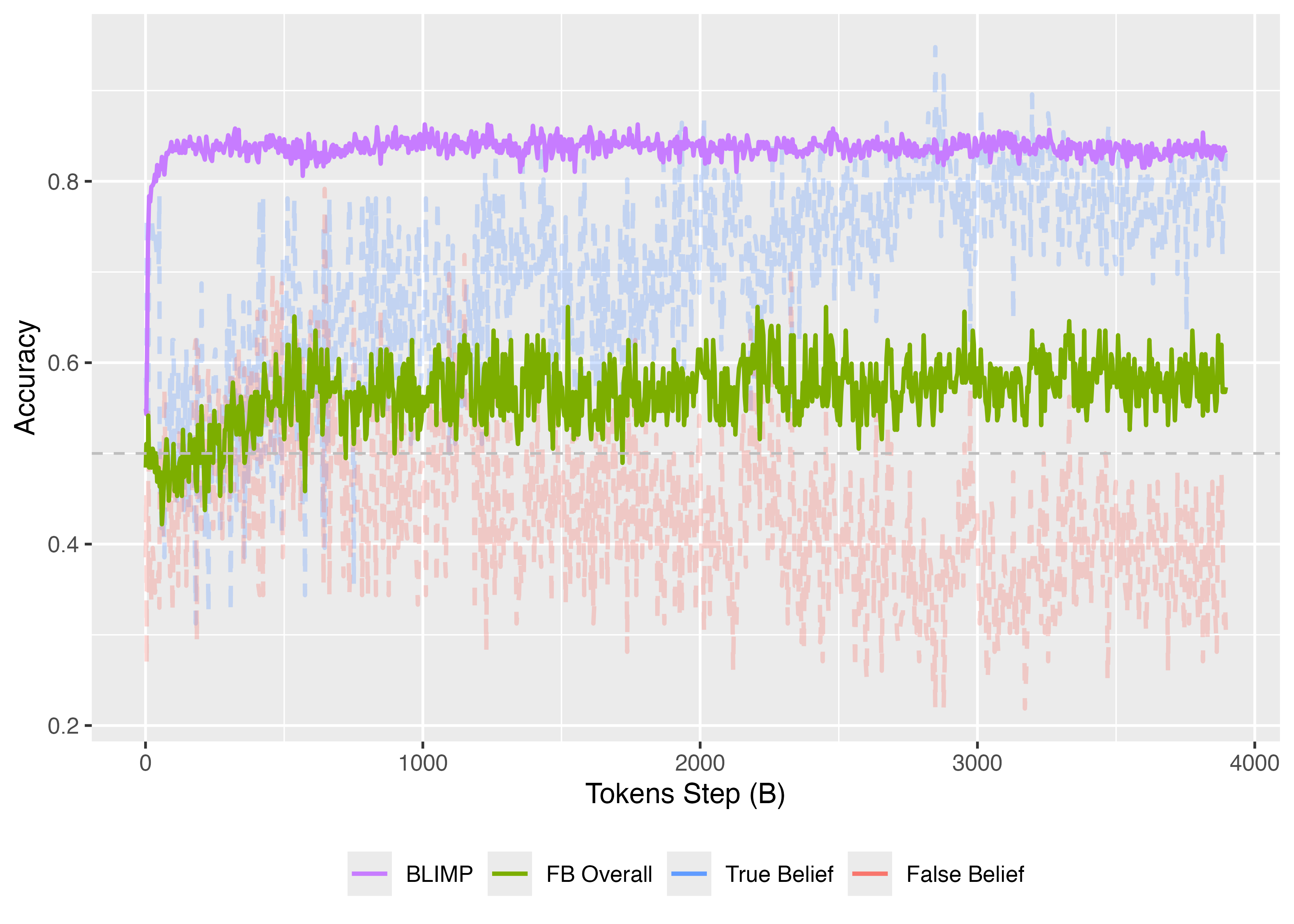}
    \caption{BLIMP vs FB task performance for OLMo 2 (7B). Solid lines show aggregated benchmark performance, whereas the dotted, lighter colours show FB performance split on knowledge state.}
    \label{fig:blimp}
\end{figure}

We also compare how quickly OLMo 2 acquires FBT capability, a \emph{functional} linguistic capability, with general \emph{formal} linguistic skill, measured through BLIMP accuracy, in Figure~\ref{fig:blimp}. 
The BLIMP benchmark contains samples for 67 linguistic phenomena, each of which isolates and tests a particular capability in syntax, morphology, or semantics \cite{warstadt2020blimp}. 
We randomly subsample the dataset to 10\% (100 samples per phenomenon), given the number of checkpoints we are repeating the benchmark across. 

The model rapidly (within 25B tokens) achieves $>80\%$ accuracy across all linguistic phenomena on BLIMP, whereas it takes at least 300B tokens to stabilise on above-chance performance for the FB tasks.
Further, there is no deterioration on BLIMP as the model trains, revealing a robust understanding of formal components. 
The behaviour on the FB dataset is also erratic and worsens during training for the False Belief knowledge state. 
Therefore, similar to earlier claims \cite{mahowald2024dissociating}, we conclude that functional and formal understanding should be measured separately. 
While the latter may be required for the former, we cannot conclude that simply exposing the model to additional tokens will automatically lead to mindreading capabilities that are required for the FB task.

\section{Data Leakage Analysis}
\label{app:leakage}
It may be possible that the contents from any of our evaluation benchmarks have made their way into the training data for any of the phases of the models used in our experiments \citep{ni2025training, balloccu2024leak}.
Our experiments investigate both open-weight models, as well as more completely open-source models, the main difference being access to their training data. For the open-source models, including \texttt{OLMo 2}, \texttt{OLMo 3}, and \texttt{K2-V2}, we can verify that the models did not directly observe test set samples at any point during training. Since our results mostly rely on results for the False Belief (FB) task, we focus our leakage analysis on this benchmark. 

\paragraph{Pre-training data} The aforementioned models publish their training data alongside the trained model. For OLMo 2, these are \texttt{olmo-mix-1124}\footnote{\url{https://huggingface.co/datasets/allenai/olmo-mix-1124}} for pre-training, and \texttt{dolmino-mix-1124}\footnote{\url{https://huggingface.co/datasets/allenai/dolmino-mix-1124}} for mid-training. 
For OLMo 3, these are \texttt{Dolma 3}\footnote{\url{https://huggingface.co/datasets/allenai/dolma3_pool}} for pre-, mid-, and long-context training, and various \texttt{Dolci 3} subsets for post-training. 
Each dataset is, in turn, a filtered combination of other datasets of various sources. 
Based on the document type in these sources, we can already disregard some of them as containing possible leakages, since they are very unlikely to include the benchmark data. 
For example, math web pages, math proofs, Wikipedia text, code, and academic paper content are unlikely to contain verbatim samples from the FB dataset. 
However, two sources potentially contain the dataset, as they are constructed by scraping the web freely (\textbf{CommonCrawl}) or by explicitly combining various datasets (\textbf{FLAN}). 
We discuss how we check each of these sources. 

\begin{itemize}[leftmargin=*, align=left, font=\bfseries, parsep=0pt, itemsep=1pt, topsep=1pt]
    \item[CommonCrawl] We can check the index to see whether any known links to the FB data have been included in the CommonCrawl corpora \citep{CommonCrawl}. We find seven hits for known links to the OSF archive (\url{https://osf.io/zp6q8}, \url{https://osf.io/agqwv/}, \url{https://archive.org/details/osf-registrations-agqwv-v1}, \url{https://archive.org/details/osf-registrations-zp6q8-v1}) for the FB dataset, though none of these contains the actual samples. For these pages, the crawler instead scraped the non-JS-enabled page text, in which the data is not rendered. 
    \item[FLAN data] FLAN data \citep{pmlr-v202-longpre23a} constitutes yet another mixture of many different datasets, and is included in \texttt{dolmino-mix-1124} and \texttt{Dolmo 3}. The bulk of this data stems from the 2021 iteration of FLAN, which predates the FB dataset and, therefore, cannot include it. However, the updated 2022 version might. Because this dataset is also included in the post-training dataset, we describe our checking mechanism there. 
\end{itemize}

\paragraph{Post-training data}
The post-training dataset used by OLMo 2 is \texttt{allenai/tulu-3-sft-olmo-2-mixture}\footnote{\url{https://huggingface.co/datasets/allenai/tulu-3-sft-olmo-2-mixture}}, or further filtered versions. 
As in the pre- and mid-training datasets, this dataset encapsulates many individual datasets as described in T\"{u}lu 3 \citep{lambert2024tulu}. Dolci 3, the post-training dataset for OLMo 3 (specifically, \texttt{Dolci-Think-SFT}\footnote{\url{https://huggingface.co/datasets/allenai/Dolci-Think-SFT}} and \texttt{Dolci-Instruct-SFT}\footnote{\url{https://huggingface.co/datasets/allenai/Dolci-Instruct-SFT}}), shares many common sources with the T\"{u}lu dataset, but is expanded to include reasoning traces, and pools from additional datasets to more than double its size.
Like before, we can already disregard certain sources as unlikely to contain the text samples from the FB benchmark, and avoid checking a set twice if it is included in both datasets. 
Since the resulting filtered set of samples to investigate is of a smaller order of magnitude (437K samples for OLMo 2, 3M samples for OLMo 3) than the pre-training data, checking for lexical overlap between FB data and the post-training samples is feasible. 

We check all 192 FB samples to see whether they appear verbatim in the chat interactions provided in the post-training datasets. 
We select the complete story text (passages), both attention-check questions, and the first critical questions to see whether any of them literally appear in the dataset's logs. 
Our search took roughly 45 minutes on modern hardware and resulted in zero matches. 
Therefore, we conclude that none of the FB samples is included in the post-training data, and this benchmark can be safely used as a true test of generalised ToM ability as a surrogate for social intelligence. 

\end{document}

%% file: figures/example.tex
\begin{tikzpicture}[
    node distance=1.5cm and 1cm,
    textStyle/.style={
        font=\footnotesize,
        text=textColor,
        align=center
    },
    shared block/.style={
        rectangle,
        rounded corners=4pt,
        fill=neutral,
        draw=gray!30,
        line width=1pt,
        text width=10cm,
        textStyle
    },
    branch block/.style={
        rectangle,
        rounded corners=4pt,
        fill=white,
        draw=gray!30,
        line width=1pt,
        text width=4.5cm,
        textStyle,
        minimum height=1cm
    },
    line/.style={
        draw,
        -Latex,
        thick,
        rounded corners,
        color=gray!60
    },
    label node/.style={
        font=\bfseries\scriptsize,
        fill=white,
        inner sep=2pt
    }
]
    \node[shared block] (start) {
        \textbf{Premise}\\
        Ed arrives home after a long day at work. He puts his keys in the hall and leaves his bag in the study. Seana arrives home a few minutes later.
    };


    \node[branch block, below left=1cm and -4.75cm of start] (true) {
       \textbf{Ed watches as Seana moves the keys from the hall to the study.} Afterwards, Ed goes to the bathroom.
    };

    \node[branch block, below right=1cm and -4.75cm of start] (false) {
        Afterwards, Ed goes to the bathroom. \textbf{Ed doesn't see Seana move the keys from the hall to the study.}
    };

    \node[shared block, below=of start, yshift=-2.2cm] (end) {
        When Ed gets back from the bathroom, he realizes he needs his keys.\\
        \textbf{Implicit Cue:} ``Ed goes to get the keys from the... (\emph{study / hall})''
        \textbf{Explicit Cue:} ``Ed thinks the keys are in the... (\emph{study / hall})''
    };

    \draw[line, color=trueColor] (start.south) -- +(0,-0.5) -| node[pos=0.25, label node, text=trueColor] {TRUE BELIEF} (true.north);

    \draw[line, color=falseColor] (start.south) -- +(0,-0.5) -| node[pos=0.25, label node, text=falseColor] {FALSE BELIEF} (false.north);

    \draw[line, color=trueColor] (true.south) |- +(0,-0.5) -- (end.north);
    \draw[line, color=falseColor] (false.south) |- +(0,-0.5) -- (end.north);
\end{tikzpicture}